# Optimal Design of Fuzzy Based Power System Stabilizer Self Tuned by Robust Search Algorithm

M.Mary Linda, Dr.N.Kesavan Nair

**Abstract**— In the interconnected power system network, instability problems are caused mainly by the low frequency oscillations of 0.2 to 2.5 Hz. The supplementary control signal in addition with AVR and high gain excitation systems are provided by means of Power System Stabilizer (PSS). Conventional power system stabilizers provide effective damping only on a particular operating point. But fuzzy based PSS provides good damping for a wide range of operating points. The bottlenecks faced in designing a fuzzy logic controller can be minimized by using appropriate optimization techniques like Genetic Algorithm, Particle Swam Optimization, Ant Colony Optimization etc. In this paper the membership functions of FLC are optimized by the new breed optimization technique called Genetic Algorithm. This design methodology is implemented on a Single Machine Infinite Bus (SMIB) system. Simulation results on SMIB show the effectiveness and robustness of the proposed PSS over a wide range of operating conditions and system configurations.

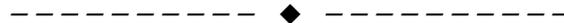

## 1 INTRODUCTION

INSTABILITY problems are caused mainly by the low frequency oscillations developed in the interconnected power system network. The major causes for this poor low frequency oscillations are sudden load changes, switching surges, both internal and external faults, transient disturbances etc., Damping of these oscillations in interconnected power system is essential for secure and stable operation of the system [2]. High gain excitation system in addition with AVR will damp out the oscillations of low frequency (0.2 to 2.5 Hz), but considerably it is favour only for the improvement of steady state stability rather than dynamic stability. Thus an auxiliary signal in addition with the excitation system and AVR will greatly improve the system dynamic stability which is termed as power system stabilizer.

Conventional power system stabilizers are designed based on eigen value analysis which utilizes two basic tuning techniques phase compensation and root locus [1]. Phase compensation is widely used and compensates for the phase lags by providing a damping torque component. Root locus involves shifting of eigen values related to the power system modes of oscillation by shifting the poles and zeros of the stabilizer [1]. CPSS are designed based on linearized theory. But as we know power systems are non-linear and are very complex. So the parameter of the CPSS is ineffective for various operating points. Also there is no interaction between the CPSS stabilizer parameters as time varies. This result in degrada-tion of the performance of the stabilizer. Later a novel approach called Fuzzy Logic has been proposed to design the PSS. Exact mathematical model is not required in designing a FLC. FLPSS is designed in the time domain wheras CPSS is designed in frequency domain [16]. This approach when compared to classical models is easy to implement and does not need a rigorous mathematical model of the system. A fuzzy logic power system stabilizer performs well and produces positive results for a wide range of operating conditions. Though fuzzy logic approach enhances the dynamic stability it also has some bottlenecks like generation of membership functions, creation of rules and choice of scaling factors which is done by trial and error method [9]. This made the design procedure a laborious one and thus became a time consuming task. Incorporation of GA in fuzzy logic power system stabilizer design will significantly reduce the time consumed in the design procedure.

On the other hand GA is a search algorithm rooted in the mechanics of natural selection and natural genetics and is used in various power system problems [10]. The main theme of GA is robustness, the balance between efficiency and efficacy necessary for survival in many different environments. GA provides an alternative to traditional optimization techniques by using directed random searches to locate optimal solutions in complex power system problems [15]. Thus the performance of fuzzy logic based power system stabilizer can be significantly enhanced by operating genetic based learning mechanism. This paper deals with the improved and novel design approach for single machine infinite bus system where the parameters are tuned using this balanced optimization technique. This is also based on the optimization criteria integral of SMSE (Sum of Mean Squared Error) [7].

―――――――――――――――――――

- M.Mary Linda is with the Department of EEE Ponjesly College of Engineering, Nagercoil and also the Research Scholar of Anna University, Chennai.
- Dr.N.Kesavan Nair is with the Department of EEE, Noorul Islam College of Engineering, Kumaracoil.

## 2 SYSTEM MODELING

Modeling is the method of developing mathematical equations for the system parameters. The basic modeling is the classical model for the generator. To this basic model the effect of synchronous machine field circuit dynamics and excitation system is added to frame the complete system block diagram as shown in Fig.1 [4].

## 3 POWER SYSTEM STABILIZER

The basic function of power system stabilizer is to add damping to the generator rotor oscillations by controlling its excitation using auxiliary stabilizing signals. To provide damping, the stabilizer produces a component of electrical torque in phase with the rotor speed deviations. The basic components in PSS are (i) Gain (ii) Washout and (iii) Phase compensation. Phase lead network provide compensation for the phase lag between the exciter input and the generator electrical torque output over the frequency range of 0.2 to 2.5 Hz. The signal washout block serves as a high pass filter with the time constant Tw. $K_{stab}$ is the stabilizer gain that determines the amount of damping introduced by the power system stabilizer [4].

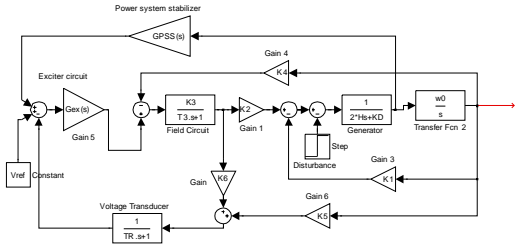

Fig. 1. Block Diagram of SMIB system

### 3.1 Conventional power system stabilizer

Conventional power system stabilizers are designed based on linearized theory and can damp out the oscillations effectively only in a particular operating point. CPSS are designed based on eigen value analysis which utilizes phase compensation (lead compensation) as its basic tuning technique. Phase compensation is widely used as it compensates for the phase lags by providing damping torque component. Since power systems are nonlinear, CPSS are ineffective for a wide range of operating conditions. This is also a time consuming process and there is no interaction between the stabilizer parameters as time varies.

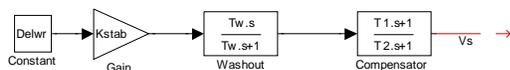

Fig. 2. Block Diagram of Power System Stabilizer

### 3.2 Fuzzy based power system stabilizer

Controller that uses fuzzy logic as its novel approach is called Fuzzy Logic Controller (FLC).In the design of FLC two inputs are considered, change in speed deviation and derivation of change in speed deviation. Seven linguistic variables are considered for each input signal which leads to the creation of 7 x7 rule table. Control signal is produced after the different steps like fuzzification, rules creation and defuzzification. Though it provides good damping it has some bottlenecks like generation of membership function, creation of rules and choice of scaling factors which is done by trial and error method [9]. To overcome these drawbacks fuzzy logic controller can be automatically tuned using the advanced optimization techniques like Genetic Algorithm, Particle Swam Optimization, and Ant Colony Optimization etc.

## 4 GENETIC ALGORITHM

A GA is an exploratory procedure that is able to locate near optimal solutions to complex problems. It maintains a set of trial solutions often called as individuals and forces them to evolve towards an acceptable solution. Generally GA's are based on two assumptions [6].
(i) An individual's fitness is an accurate measure of its relative ability to solve the problems
(ii) That combining individuals will enable the formation of improved off spring.
 GA is a four step process involving evaluation, reproduction, recombination and mutation.

### i) Representation

Encoding of chromosomes (also called as individuals) is termed as representation which is also referred as chromosomes. In common practice chromosomes are represented by group of strings.

### ii) Evaluation

This is the first step in each generation which is done in current chromosomes. Each chromosome in the population is decoded and evaluated on how well it solves the problem.  This fitness measure is used in the next step to determine how many off springs will be generated from any particular chromosome.  Basically in evaluation process individuals are selected from the populations for reproduction based on more highly fit individuals.  These selected individuals form pairs called parents.  Selection is the key operation on GA that ensures the survival of the fittest. Commonly used selection methods are ratioing and ranking

### iii) Reproduction

It is the process of copying individual strings according to their fitness function.Copying strings according to their fitness values means that strings with the higher value have a higher probability of contributing one or more off springs in the generations.

### iv) Recombination

Reproduction creates a population whose members currently best solve the problem.  However many of the chromosomes are identical and some are different from the previous generations. Reproduction produces multiple copies of existing chromosomes from the population and produces a new chromosome that maintains many of the features of the previous generation. The most common method for recombination is crossover.

**v) Crossover**

It is a genetic operator used to vary the chromosomes from one generation to the next. It combines portions of two parents or individuals to create two new individuals called off spring which inherits the features of the parents.

**vi) Mutation**

It is the last step is creating a new generation. It is termed as the genetic operator used to maintain genetic diversity from one generation of a population of chromosomes to the next. It is also referred as the injection of new information into the population [6]. The basic mutation operators that are commonly used are flipping, reversing and interchanging .Reproduction and crossover together give GA most of their power whereas mutation helps to increase the searching power. Mutation as it does in nature, takes place very rarely on the order of once in a thousand bit string location.

### 4.1 Simple GA algorithm:

1. Create a Population of strings or chromosomes
2. Select two individuals from the population
3. Cross the two individuals
4. Generate two new individuals
5. Mutate characters in the new individuals
6. Select more highly fit individuals
7. Thus obtain the population in an unordered set
8. Repeat the process of selection crossover and mutation till the entries of the new generation are filled
9. Now the old generation is discarded
10. New generations are generated till some stopping criteria are involved [13].

### 4.2 Termination

GA process is terminated under the following criteria.

(i) Fixed number of generation
(ii) If all individuals in the population converge to the same string
(iii) If all the minimum criteria are satisfied
(iv) If these is no improvement in fitness values that are found after a given number [13].

### 4.3. Optimization of membership function using GA

GA as first described can be used to optimize the membership function. Optimization is relatively the process of adjusting the base value of membership function. Given some functional mapping for a system some membership functions and their shapes are assumed for the various fuzzy variables defined for a problem. These M.F's are then coded as bit strings that are then concatenated. An evaluation (fitness) function is used to evaluate the fitness of each set of parameters that defuse the functional mapping (membership function) Let we assume that each M.F has the shape of a right angle and 6-bit binary strings are used to define the base of M.F .These strings are then concatenated to give us a 24 bit string. Let we start with an initial population of four strings which are decoded to the binary values of the variables. The binary values are then mapped to decimal values for the fuzzy variables. To determine the fitness of the combination of M.F's in each of the strings we have to measure the square of the errors that are produced in estimating the value of the outputs for the given inputs. Then the sum of the squared errors is subtracted from a standard value to convert the fitness function from a minimization problem. Thus the fitness value is obtained. Then we calculate the sum of all the fitness values and its average fitness in the generation is used to determine the relative fitness of the strings in the generation and has been used for the acceptability of a string propagating into the next generation. After this crossover and mutation operations are done, this increases the fitness of the best string in the second generation [11].

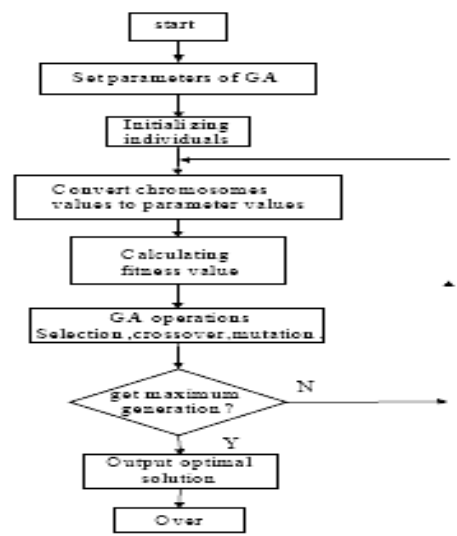

**Fig.3. Flow Chart for parameter optimization using Genetic Algorithm [14]**

Thus we obtain a M.F which is overlapped that shows a very desirable property of M.F. This process of generating and evaluating strings is continued until we get a convergence to the solution within a generation i.e., we get the M.F's with the best fitness value [11].The main parameters to be optimized using GA are PSS gain $K_{stab}$ and phase lead compensator time constant [12].

Minimization of the integral of sum of the squares of the speed deviation signal $\Delta\omega$ is considered as the objective function the. [12]

## 5 TUNING RESULTS AND DISCUSSION

The dynamic performance of one machine system is obtained for the following loading conditions
1. Light loading condition (P=0.4 p.u; Q=0.5 p.u; Large step disturbance=0.1p.u)
2. Nominal Operating Condition (P=1.0 p.u; Q=0.015 p.u; small step disturbance=0.01p.u)
3. Heavy loading condition (P=1.25 p.u; Q=0.25 p.u; small step disturbance=0.01p.u)

A step perturbation is given and the dynamic performance of the system with the improved robust algorithm based PSS is compared with the conventional PSS and without PSS.

## 5.1 Light loading condition

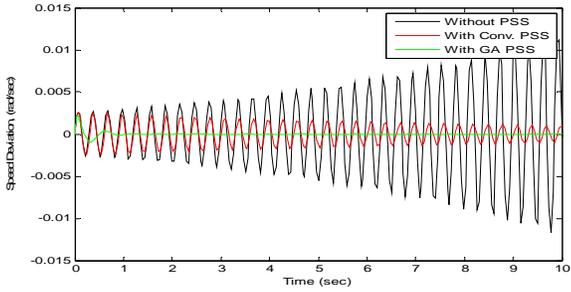

Fig.4a. Speed Deviation under Light loading condition

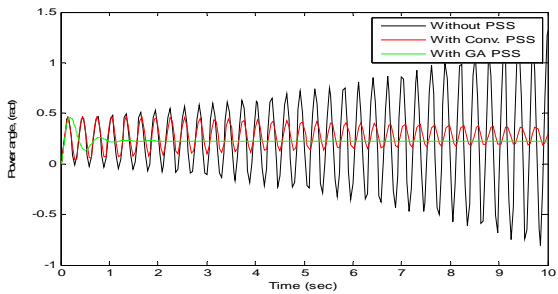

Fig.4b. Power Angle Deviation under Light loading condition

## 5.2 Nominal loading Condition

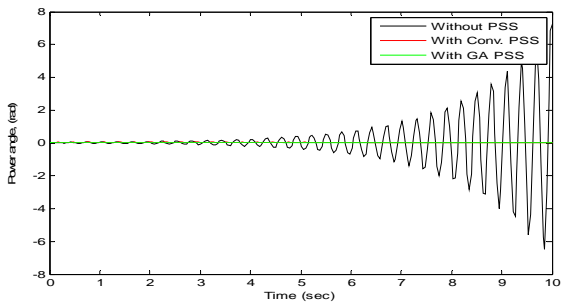

Fig.5a. Speed Deviation under nominal loading condition

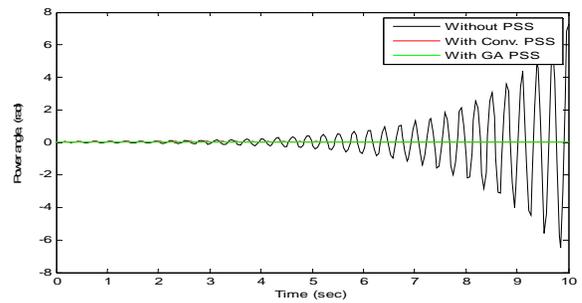

Fig.5b. Power angle deviation under nominal loading condition

## 5.3 Heavy Loading Condition

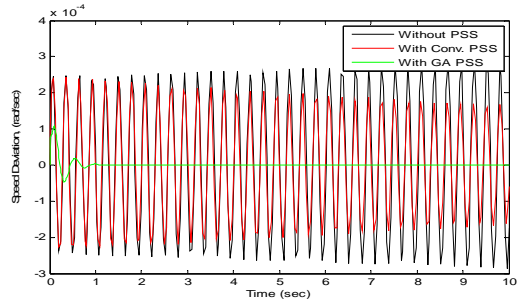

Fig.6a. Speed Deviation under heavy loading condition

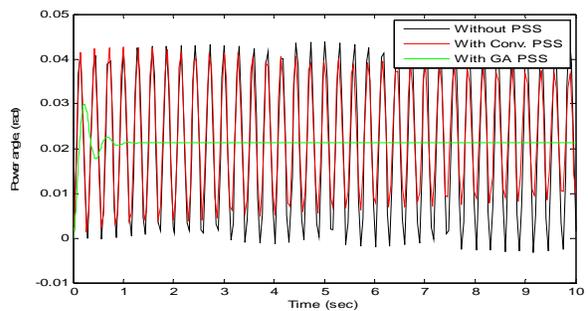

Fig.6b. Power Angle Deviation under heavy loading condition

**Case 1:**
Simulated figures 4a and 4b shows the comparison of dynamic performances under a light loading condition. In this case, improved GA based PSS provides very good damping (Damping time constant is 1.5 sec) as compared to both with conventional PSS and without PSS.

**Case 2:**
Simulated figures 5a and 5b shows the comparison of dynamic performances under a nominal loading condition. In this case, both improved GA based PSS and conventional PSS provides good damping (GAPSS is overlapped by CPSS) as compared to without PSS.

**Case 3:**
Simulated figures 6a and 6b shows the comparison of dynamic performances under a heavy loading condition. In this case, GA based PSS provides very good damping (Damping time constant is 1.1 sec) as compared to both with conventional PSS and without PSS. Thus the improved robust algorithm based PSS continues to perform well for all operating loads and hence has a higher level of robustness.

## 6 CONCLUSION

An intelligent approach for designing fuzzy based power system stabilizer for a single machine infinite bus system has been presented. Simulation of the response to small and large disturbances has demonstrated the effectiveness of this robust algorithm. It is shown that the proposed robust optimization provides good damping characteristics and enhances the dynamic stability of the system. Its level of robustness to system load variations is better than conventionally tuned PSS.

As the number of energy suppliers connected to the network increases this research can be directed to develop systematic methods of design and analysis for fuzzy based stabilization control like Particle Swam optimization, Ant Colony Optimization etc. The simulation results of various parameter optimization techniques can be compared for multimachine system and its effectiveness can be studied.

## APPENDIX

The generator and exciter parameters all in per unit of a SMIB system are as follows [5]:
Generator details:
M=9.26    $T_{do1}$=7.76    D=0    $X_d$=0.973
$X_{d1}$=0.190    $X_q$=0.550
Exciter details:
Ka=50    Ta=0.05    $K_f$=0.025    $T_f$=1
Line and load details:
R=-0.034    X=0.997    G=0.249
B=0.262    f=60
Initial conditions:
$P_{e0}$=1.0    $Q_{e0}$=0.015    $V_{t0}$=1.05

**M.Mary Linda** received her B.E(Electrical and Electronics Engg;) degree from Noorul Islam College of Engineering, Kumaracoil in the year 2002 and her M.E (Power Systems) from Arulmigu Kalasalingam College of Engineering, Srivilliputhur in the year 2005. She has achieved the Gold Medal in P.G Degree from Anna University, Chennai in the year 2005. She has the teaching experience of six years. Her current research interests are AI techniques and power system stability.

**Dr.N.Kesavan Nair** B.Sc. (Engg.), M.Sc. (Engg.), Ph.D. (I.I.T Kharagpur) is a retired professor from Govt. Engineering College, Trivandrum and Trichur. He has to his credit a few publications in his area at national and international level.Since 1997 he has been working as visiting Professor of Electrical Engg. in various engineering colleges.